\documentclass[a4paper,twoside]{article}

\usepackage{epsfig}
\usepackage{subcaption}
\usepackage{calc}
\usepackage{amssymb}
\usepackage{amstext}
\usepackage{amsmath}
\usepackage{amsthm}
\usepackage{multicol}
\usepackage{pslatex}
\usepackage{apalike}
\usepackage{SCITEPRESS}     
\usepackage{lipsum}
\usepackage{mathtools}
\usepackage{listings}

\usepackage{algorithm}
\usepackage{algpseudocode}
\usepackage{varwidth}

\usepackage{tikz}
\usetikzlibrary{fit,shapes, positioning}

\theoremstyle{theorem}
\newtheorem{mydef}{Definition}

\theoremstyle{definition}

\begin{document}

\title{Multi-Agent Systems based on Contextual Defeasible Logic considering Focus}

\author{\authorname{Helio H. L. C. Monte-Alto\sup{1}, Mariela Morveli-Espinoza\sup{1} and Cesar A. Tacla\sup{1}}
\affiliation{\sup{1}Programa de Pós-Graduação em Engenharia Elétrica e Informática Industrial, Universidade Tecnológica Federal do Paraná, Curitiba - PR, Brazil}
\email{heliohenrique3@gmail.com, morveli.espinoza@gmail.com, tacla@utfpr.edu.br}
}


\abstract{In this paper, we extend previous work on distributed reasoning using Contextual Defeasible Logic (CDL), which enables decentralised distributed reasoning based on a distributed knowledge base, such that the knowledge from different knowledge bases may conflict with each other. However, there are many use case scenarios that are not possible to represent in this model. One kind of such scenarios are the ones that require that agents share and reason with relevant knowledge when issuing a query to others. Another kind of scenarios are those in which the bindings among the agents (defined by means of mapping rules) are not static, such as in knowledge-intensive and dynamic environments. This work presents a multi-agent model based on CDL that not only allows agents to reason with their local knowledge bases and mapping rules, but also allows agents to reason about relevant knowledge (focus) -- which are not known by the agents \textit{a priori} -- in the context of a specific query. We present a use case scenario, some formalisations of the model proposed, and an initial implementation based on the BDI (\textit{Belief-Desire-Intention}) agent model.}

\onecolumn \maketitle \normalsize \setcounter{footnote}{0} \vfill

\section{\uppercase{Introduction}}
\label{sec:introduction}

Contextual Defeasible Logic (CDL), introduced by \cite{bikakis2010}, was proposed as a contextual reasoning model that addresses the requirements of intelligent environments, such as Ambient Intelligence systems, in which various agents process, change and share available context information. Such information is often imperfect and may pose conflicts among agents when reasoning about them. CDL bases its notions on the Multi-Context Systems (MCS) paradigm, according to which local knowledge of agents is encoded in rule theories (which we will call \textit{belief bases} in this work\footnote{In the original work on CDL, and also in other works on MCS, such rule theories are called \textit{contexts}. However, in order to avoid confusion about what is \textit{context} and \textit{context information}, which is a relevant conceptualisation in this work, we will use the term \textit{belief base} to mean the local knowledge held by an agent}), and information flow between agents is achieved through mapping rules that associate concepts used in different belief bases. To resolve potential conflicts that may arise from the interaction of agents through their mappings (\textit{global conflicts}), a preference order on the known agents is used, in order to express the confidence that an agent has in the knowledge imported by other agents. Conflicts are resolved by means of defeasible argumentation semantics. 

The main aim of CDL (and of works on MCS in general) is to accomplish \textit{contextual reasoning} in a distributed setting. \cite{benerecetti2000} defines that contextual reasoning corresponds to operating on three fundamental dimensions along which a context dependent representation may vary: \textit{partiality}, namely with the portion of the world which is taken into account; \textit{approximation}, namely with the level of detail at which a portion of the world is represented; and \textit{perspective}, namely with the point of view from which the world is observed. CDL implements mainly the \textit{partiality} dimension, i.e. it enables reasoning with incomplete and imperfect information spread among different agents. However, the \textit{approximation} dimension is not well studied by \cite{bikakis2010}. The idea of approximation is related to the idea of \textit{focus}: the closer one is to an object of interest, the more it is aware of what is relevant about that object. 

Therefore, although the CDL model is able to perform decentralised distributed reasoning and handle the conflicts between knowledge from different agents, there is a lack of capabilities to allow agents to reason about a topic or object that may not be known \textit{a priori} by all agents. For example, when an agent (which we will call \textit{starter agent}) wants to reason about a recent perception about some object found in the environment, and therefore needs help from other agents to achieve some conclusion (e.g. a property of the object perceived), it would be necessary to share relevant information about such object to enable other agents to be aware of the object's properties, and then effectively reason about it. Intuitively, the agent has to ask something like: ``Do someone knows if this is true given these properties that I have just found?''. On the other hand, the original CDL approach is able to answer only the question: ``Do someone knows if this is true?''.

The relevant context information about the object that the starter agent wants to reason about is called, in this work, \textit{focus}, in the sense that it concerns a set of rules and/or facts that is relevant at a given moment for a particular reasoning context. All the agents involved in the reasoning must be aware of this focus in order to reach the most reasonable conclusion 
in a fully decentralised manner. The scenario presented in Section \ref{sec:scenario} illustrates such problem and how it could be potentially solved. We are also concerned with the way that agents can reason in a scenario where multiple \textit{foci} from different interactions between agents should be considered. Agents should be able to 
reason correctly considering multiple contexts at the same time.

Furthermore, this work extends CDL by allowing mapping rules that reference knowledge held by any known agent, instead of only enabling references to a specific agent. Therefore, mapping rules can also have literals in their bodies that are called \textit{schematic}, such that the actual agents that must be considered to import knowledge from must be determined in execution time. 
This enables the use of the model in dynamic environments, where agents come in and out of the system at any time, and on knowledge-intensive environments where agents have their belief bases updated and revised constantly. This contribution will not be covered in detail in this paper, but will be used and explained briefly in the scenario presented in Section \ref{sec:scenario} due to the difficulty of representing it without \textit{schematic} terms.

This paper is divided in four more sections. Section \ref{sec:scenario} presents a scenario which illustrates the problem, as well as high-level description of the desirable behaviour for the agents in the scenario. Section \ref{sec:formalisation} presents the formalisation, which includes: the definition of the elements in the agents' architecture, the definition of the types of rules used to represent knowledge; and the description of the arguments construction. Section \ref{sec:implementation} presents an initial implementation using the multi-agent paradigm. Finally, Section \ref{sec:conclusions} presents related works, conclusions and future works.

\section{\uppercase{Scenario}}
\label{sec:scenario}

This scenario, first introduced by \cite{bikakis2009tese} in his thesis, takes place in an Ambient Intelligence environment of mushroom hunters, who collect mushrooms in a natural park in North America. The hunters carry mobile devices, which they use to communicate with each other through a wireless network in order to share their knowledge on edible and non-edible mushrooms. 

The goal of mushroom hunting for every agent is to collect edible mushrooms, but typically people do not know every species and family of mushrooms in detail. Let's suppose an agent Alice that has some knowledge on some specific species, such as the \textit{Death Cap} and the \textit{Caesar's Mushroom}. She knows that the former is poisonous and the latter is not. She is also willing to believe that a mushroom is edible or not if any other known agent believes the same. This knowledge is represented by mapping rules with schematic terms, meaning, for example, that she is willing to accept that the mushroom is edible if some other agent affirms it. Of course, it is possible that different agents state contradictory information (e.g. $edible(m)$ and $\neg edible(m)$), which will be resolved by means of an argumentation based conflict resolution. Thus, her rules can be denoted as:

\begin{equation*}
\begin{split}
r^{@}_{A1}: \neg edible(M)_A \Leftarrow mushroom(M)_A, \\
death\_cap(M)_@ \\
r^{@}_{A2}: edible(M)_A \Leftarrow mushroom(M)_A, \\
caesar\_mushroom(M)_@ \\
r^{@}_{A3}: edible(M) \Leftarrow  edible(M)_@ \\
r^{@}_{A4}: \neg edible(M)  \Leftarrow \neg edible(M)_@ \\
\end{split}
\end{equation*}

At some moment, she finds a mushroom with the following characteristics: It has a stem base featuring a fairly prominent sack that encloses the bottom of the stem (\textit{volva}), a pale brownish cap with patches, the margin of the cup is prominently lined, and the mushroom does not have a ring (\textit{annulus}). When performing her local reasoning process, using the just perceived facts together with her current belief base, Alice is not able to reach a conclusion about the edibleness of the mushroom. Thus, she has to query other agents known by her, which are Bob, Catherine, Dennis and Eric. Alice has different levels of confidence in them: she trusts more in Eric, followed by Catherine, and then Bob and Dennis. Thus, the preference order of Alice is denoted as $T_A = [E, C, B, D]$.

In order to enable them to reason effectively about the mushroom perceived, Alice has to share some information about it to them, otherwise they would not be able to reach any reasonable conclusion. Thus, she sends a query (denoted here by \textit{edible(m1)?}, where \textit{m1} represents the mushroom found), and together with the query she sends the facts (in the form of rules without bodies) that describe the properties of the mushroom. Besides that, an identification for the context of the query is sent ($C_Q = \alpha$), in order to enable a traceability of the rules exchanged between agents during the reasoning process for the query, and to avoid confusion among reasoning processes that may be executing simultaneously by the same agents. Note that the names of the rules are tagged with the same query context identification. The complete content of the query is formalised as follows:

\begin{equation*}
\begin{split}
C_Q: \alpha \\
p: edible(m1)? \\
r^\mathcal{F}_{\alpha_{11}}:  mushroom(m1) \Leftarrow \\
r^\mathcal{F}_{\alpha_{12}}:  has\_volva(m1) \Leftarrow \\
r^\mathcal{F}_{\alpha_{13}}: pale\_brownish\_cap(m1)  \Leftarrow\\
r^\mathcal{F}_{\alpha_{14}}:  patches(m1) \Leftarrow \\
r^\mathcal{F}_{\alpha_{15}}:  cup\_margin\_lined(m1) \Leftarrow \\
r^\mathcal{F}_{\alpha_{16}}:  \neg has\_annulus(m1) \Leftarrow
\end{split}
\end{equation*}

The agents Bob, Catherine, Dennis and Eric receive this query, then each one performs their reasoning process using the union of their local belief bases with the focus rules received from Alice. Bob has the following rules in his belief base:

\[ r^{d}_{B1}: \neg edible(M)_B \Leftarrow mushroom(M)_B, has\_volva(M)_B \]
\[ r^{@}_{B1}: death\_cap(M)_B \Leftarrow death\_cap(M)_A  \]

which means that he concludes that an object is not edible if it is a mushroom and it has \textit{volva}. He also knows that Alice knows something about \textit{death caps} which is represented by the mapping rule $r^{m}_{B1}$.


Catherine has the following rule in her belief base:
\begin{equation*}
\begin{split}
r^{@}_{C1}: edible(M)_C  \Leftarrow mushroom(M)_C, \\springtime\_amanita(M)_@
\end{split}
\end{equation*}
which means that she believes that an object is edible if it is a mushroom and it is of the species \textit{springtime amanita}. However, Catherine is unable to describe a mushroom of this species, thus the rule is a mapping rule with a schematic term, meaning that she is willing to accept that the mushroom is a \textit{springtime amanita} if someone affirms it. 

Dennis has the following rule:

\begin{equation*}
\begin{split}
r^{@}_{D1}: \neg edible(M)_D \Leftarrow mushroom(M)_D, amanita(M)_@
\end{split}
\end{equation*}

which means that he believes that an object is not edible if it is an amanita. However, Dennis does not know anything about amanita, thus a schematic term is used in the body of the rule.

Finally, Eric's rules can be described as follows:

\begin{equation*}
\centering
\begin{split}
    r^{d}_{E1}: springtime\_amanita(M)_E  \\ \Leftarrow mushroom(M)_E, has\_volva(M)_E, \\ pale\_brownish\_cap(M)_E, patches(M)_E, \\ cup\_margin\_lined(M)_E, \neg has\_annulus(M)_E \\
    r^{ck}_{1}: amanita(M) \leftarrow springtime\_amanita(M) \\
\end{split}
\end{equation*}

which means that he concludes that a mushroom is a \textit{springtime amanita} if it has some properties, which are the same properties perceived by Alice. He also holds a common knowledge believed by all agents that \textit{springtime amanitas} are a kind of amanita, which is represented by the rule $r^{ck}_{1}$.

Given this state of things and the query sent by Alice to each one of the other agents described, the following distributed reasoning process, described in a high-level language, occurs. A sequence diagram, already considering some implementation nomenclature, is also presented in Figure \ref{fig:seqdiagram} at the end of the paper. Note that the order of the messages exchanged between the agents not necessarily must be the same as described, since the messages are exchanged in an asynchronous manner. 

\begin{enumerate}
    \item Bob receives the query, and by considering his knowledge (rules) together with the focus rules, concludes $\neg edible(m1)_B$, answering it to Alice;
    \item Alice receives Bob's answer, which activates her rule $r^@_{A4}$. Internally, Alice keeps this result and waits for the answers from the other agents;
    \item Catherine receives the query and is not able to reach any conclusions locally using her rules. Thus, the mapping rule $r^@_{C1}$ is used, resulting in Catherine sending a query ``$springtime\_amanita(m1)?$'' to every known agent (except Alice, who is the originator of the context of the query), sending also the focus rules originated from Alice's query;
    \item Bob and Dennis receive the query from Catherine, but they do not have any rule whose term in the body matches $springtime\_amanita(m1)$, thus they answer $undefined$ to Catherine. However, Eric, using the focus rules together with his rule $r^{d}_{E1}$, concludes that $springtime\_amanita(m1)$ is true, then answering Catherine with this truth value;
    \item Catherine receives the answer from Eric. As there is no other rule to consider that opposes to $springtime\_amanita(m1)$, she answers $edible(m1)$ to Alice;
    \item Alice receives the answer from Catherine and keeps it until the other agents queried for $edible(m1)?$ return their answers;
    \item Dennis receives the query and, by means of his rule $r^@_{D1}$ and the focus rules, concludes $\neg edible(m1)$, returning this answer to Alice;
    \item Alice receives the answer from Dennis. Alice also receives the answer from Eric, who answers $undefined$ given that he does not have a rule with head $edible(M)$ or $\neg edible(M)$. As there is already a $\neg edible$ answer (given by Bob), Alice has to decide which one to consider. Using the preference order, she chooses the answer from Bob, in who she trusts more than in Dennis.  Finally, given that she received two conflicting answers ($edible(m1)$ and $\neg edible(m1)$), she again applies her preference order. Since she trusts more in Catherine than in Bob, she chooses the answer $edible(m1)$, concluding the reasoning process.

\end{enumerate}

Given this high-level description of the scenario and the desirable behaviour and outcome, the next section formalises the framework.

\section{\uppercase{Formalisation}}
\label{sec:formalisation}

\subsection{Multi-Agent and Agents' Internal Architecture}

A Contextual Defeasible Multi-Agent System (CDMAS) is defined as $M = \{A, R_{CK}, C_Q\}$, such that $A = \{A_1, A_2,... A_n\}$ is the set of \textit{agents} currently situated in the environment, $R_{CK}$ is a set of \textit{common knowledge rules}, and $C_Q$ is a set of \textit{query contexts}. 

$R_{CK}$ represents knowledge that must be considered by every agent. Such kind of knowledge can also be called \textit{domain or background knowledge} \cite{homola2015}, which differs from \textit{contextual knowledge}, which can be thought as knowledge held by one or more specific agents and which depends on the current situation of the system. 

$C_Q$ is a set of symbols $\{\alpha, \beta, ..., \omega\}$ used to identify uniquely different query contexts that may be in progress simultaneously. Such identifications are used to tag knowledge exchanged between agents during the distributed reasoning process in order to allow agents to remind what rules can be used for specific queries. 

An agent $A_i$ is defined as a tuple of the form $(B_i, A^K_i, T_i)$, such that: $B_i$ is the belief base of the agent; $A^K_i$ is a set of agents known by $A_i$; and $T_i$ is a preference relation as a total order relation over $A_{K_i}$;

$B_i$ is defined as a tuple of the form $(V_i, R_i, R_{\mathcal{F}_i})$, where $V_i$ is the vocabulary of the agent, i.e. the literals used in its rules, $R_i$ is a finite set of rules representing the consolidated beliefs of the agent and $R_{\mathcal{F}_i}$ is a set of rules representing the current focus originated from the agent, such that these rules may or not be in $R_i$. 

In expressing the proof theory (which is based on defeasible argumentation, as explained in Section \ref{sec:arg_construction}) we consider only propositional-like rules. Rules containing free variables are interpreted as the set of their variable-free instances, thus assuming that a unification algorithm is used during the reasoning process, resulting in propositional rules. Therefore, the terms in the definition of the rules can be represented as predicates, but at execution time such rules will be instantiated with propositional literals.

$R_i$ consists of three subsets of rules, described as follows:

     \textbf{Local strict rules} ($R_{l_i}$), of the form
        \[ r^l_i: a^n_i \leftarrow a^1_i, a^2_i, ...a^{n-1}_i   \]
        
        Note that the literals are tagged with the identification of the agent ($i$) in order to indicate that such literals are defined locally (i.e. $a^k_i \in V_i$ for $1 \leq k \leq n$). Local static rules are interpreted by classical logic: whenever the literals in the body are logical consequences of the local theory, then the literal in the head ($a^n_i$) is also a logical consequence. A strict rule with empty body denotes factual knowledge without uncertainty.
        
      \textbf{Local defeasible rules} ($R_{d_i}$), of the form
        \[ r^d_i:  b^n_i \Leftarrow  b^1_i, b^2_i,...b^{n-1}_i  \]
        
        Such rules are used to express uncertainty, in the sense that a defeasible rule ($r^d_i$) cannot be applied to support a conclusion ($b^n_i$) if there is adequate (not inferior) contrary evidence. A local defeasible rule with empty body denotes factual knowledge that may be defeated.
        
     \textbf{Mapping defeasible rules}, of the form
        \[  r^@_i: c^n_i \Leftarrow c^1_i, c^2_j,..., c^k_@, ... c^{n-1}_k  \]

        A defeasible mapping rule has at least one term in the body that is defined by another agent. This means that, in order to state that the head ($c^n_i$) is a logical consequence of the system, it is necessary that the members of the body defined externally (e.g. $c^2_j$, which is defined by $A_j$) must be a logical consequence of the system, as well as the members of the body defined locally. 
        
        There is also the case when the agent that defines a literal is undefined, which is represented by a \textit{schematic term} (e.g. $a^k_@$). In this case, every known agent $A_y \in A^k_i$ must be queried in order to find out if the literal is a logical consequence of the system or not. 
        
        The intuition of a defeasible mapping rule denoted by $r^@_i: b_i \Leftarrow a_j$ is the following: \textit{``If the agent $A_j$ asserts $a$, then $A_i$ considers it a valid premise to conclude $b$ if there is not any adequate contrary evidence''}. The intuition of a defeasible mapping rule whose body consists of a single schematic term, denoted by $r^@_i: b_i \Leftarrow a_@$ is the following: \textit{``If \textbf{any} agent asserts $a$, then $A_i$ considers it a valid premise to conclude $b$ if there is not any adequate contrary evidence''}. 

The focus rules ($R_{\mathcal{F}_i}$) is also a set of rules (mainly factual ones, having an empty body) that denotes the current focus on the context. Focus rules can be used to represent the agent's current perceptions, or simply some set of facts and/or rules that compose the topic of discussion or current concern of the agent at a given moment. Additionally, focus rules are tagged with a query context $c_{i} \in C_Q$ in order to indicate which query context they belong to.

Common knowledge rules ($R_{CK}$) can be either local strict rules or local defeasible rules. It makes no sense for them to be mapping rules because this kind of knowledge is independent of the agents existing in the environment.



Finally, each agent $A_i$ defines a \textit{total preference order} $T_i$ over its known agents $A^k_i$ in order to express its confidence in the knowledge imported from an agent comparing to other agents. Therefore, this preference order has the form: 
\[ T_i = [A_k, A_l, A_m, ..., A_z]  \]

such that, for $A_i$, $A_k$ is preferable to $A_l$, $A_l$ to $A_m$, and so on. The preference order enables resolution of conflicts that may arise from the interaction among agents through their mapping rules.

\subsection{Arguments Construction and Semantics}
\label{sec:arg_construction}

This section presents how arguments are constructed in a declarative and formal manner. Each agent builds its arguments to support literals considering its own rules together with focus rules received from other agents. However, some knowledge present in mapping rules must be imported from other agents. Therefore, arguments from different agents involved in a given query context $\alpha$ must be interrelated by means of a \textit{Support Relation} ($SR_\alpha$).

\textit{Support Relations} are based on the set of all the rules considered and relevant to a given query context. Thus, for each query context $\alpha$ and for each agent $A_i$, we define $R^Q_{i\alpha}$ as the set of rules in $A_i$ tagged with $\alpha$ -- which means that such rules are those coming from focus rules exchanged during the reasoning process for a specific query. We also define the \textit{extended rule set} considered for $\alpha$ in agent $A_i$ as $R^+_{i\alpha} = R_i \cup R^Q_{i\alpha}$.

Another issue that must be considered before defining $SR_\alpha$ regards the mapping rules with schematic terms, because an SR defines proof trees whose leafs make reference to specific literals defined by other agents. In fact, a proof tree only exists if, for all literals in its leafs, there is another proof tree which has this literal as a conclusion of the tree. Therefore, we will assume the existence of a \textit{rule instantiation function} $\Lambda: 2^R \rightarrow 2^R$ (i.e., a function that takes a set of rules and gives a new set of rules), which creates a new mapping rule for every mapping rule with schematic terms, and for every possible substitution of a schematic term for a concrete term which refers to an actual literal supported by some other agent\footnote{In fact, such instantiation mechanism has some similarities to unification from first-order logic, but instead of instantiating literals we instantiate rules. If the rules are written as predicates and the literals supported by the other contexts as variables, this could be a way to define it. However, the full definition of this instantiation mechanism will be further detailed in future papers.}. The schematic rules are then excluded, remaining only local rules and mapping rules without schematic terms. We will call the set of rules (without schematic terms) that must be considered for the query context $\alpha$ in agent $A_i$ as $R^*_{i\alpha}$, such that $R^*_{i\alpha} = \Lambda(R^+_{i\alpha})$. 

\begin{mydef}
The \textit{Support Relation} $SR_\alpha$ for a query context $\alpha$ is defined as the set of triples of the form $(A_i, PT_{p_i}, p_i)$, where $PT_{p_i}$ is a proof tree for $p_i$ based on the extended set of rules (after instantiation of the rules with schematic terms) $R^*_{i\alpha}$  of $A_i$. $PT_{p_i}$ is a tree with nodes labeled by literals such that the root is labeled by $p_i$, and for every node with label $q$:

\begin{enumerate}
    \item If $q \in V_i$ and $a_1, ..., a_n$ label the children of $q$ then:
    \begin{itemize}
        \item if $\forall a_i \in \{a_1,...,a_n\}:$ $a_i \in V_i$ or $a_i = \top$ then there is a local rule $r_i \in R^*_{i\alpha}$ with body $a_1,...,a_n$ and head $q$
        \item if $\exists a_j \in \{a_1,...,a_n\}$ such that $a_j \not \in V_i$ then there is a mapping rule $r_i \in R^*_{i\alpha}$ with body $a_1,...,a_n$ and head $q$
    \end{itemize}
    \item if $q \in V_j \not = V_i$, then this is a leaf node of the tree and there is a triple of the form $(A_j, PT_q, q)$ in $SR_\alpha$
     \item If $q = \top$, then this is a leaf node of the tree.
\end{enumerate}
\end{mydef}

Arguments are defined as triples in the \textit{Support Relation}.

\begin{mydef}
An argument $\mathcal{A}$ for a literal $p_i$ in the query context $\alpha$ is a triple $(A_i, PT_{p_i}, p_i)$ in $SR_\alpha$.
\end{mydef}

Any literal labelling a node of $PT_{p_i}$ is called \textit{a conclusion} of $\mathcal{A}$. However, when we refer to \textit{the conclusion} of $\mathcal{A}$, we refer to the literal labelling the root of $PT_{p_i}$ ($p_i$). We write $r \in \mathcal{A}$ to denote that the rule $r$ is used in the proof tree of $\mathcal{A}$. The definition of subarguments (Definition 3) is based on the notion of subtrees.

\begin{mydef}
A (proper) subargument of A is every argument with a proof tree that is a (proper) subtree of the proof tree of A.
\end{mydef}




Once the arguments are constructed, we need a way to determine which arguments (and consequently literals) will be accepted as logical consequences or not. An \textit{argumentation semantics} aims to define which arguments are considered accepted and which arguments are considered rejected. As we will see, this framework also allows the idea of an argument that is neither accepted nor rejected. In fact, the argumentation semantics of CD-MAS is identical to the argumentation semantics of CDL , thus we recommend the readers to \cite{bikakis2009tese} and \cite{bikakis2010} for a complete description of the semantics, due to the paper's space limits. We will just present, in the example that follows, which arguments will be accepted and which will be rejected, briefly explaining the intuition for the decision.



\subsubsection{Argument Construction and Semantics Example}

\begin{figure*}
\centering
    \begin{tikzpicture}[ edge from parent/.style={draw,latex-}]
        \node (N1) at (-5,0) {$ed(m1)_A$}
                        child {
                            node {$ed(m1)_C$}
                        };
                
        \node[above = 0.7cm of N1,align=center,anchor=center](A1) {$A_{11}$};

       \node (N2)  [right= 0.0cm of N1] {$\neg ed(m1)_A$}  
                child {
                    node {$\neg ed(m1)_B$}
                };
                
         \node[above = 0.7cm of N2,align=center,anchor=center] (A12) {$A_{12}$};

        \node (N3) [right= 0 cm of N2] {$\neg ed(m1)_A$}  
                child {
                    node {$\neg ed(m1)_D$}
                };
         \node[above = 0.7cm of N3,align=center,anchor=center] (A12) {$A_{13}$};
    
         \node (N10) [right= 1.0 cm of N3] {$\neg ed(m1)_B$}  
                child {
                    node {$mus(m1)_B$}
                    child {
                        node {}
                    }
                }
                child {
                    node {$hv(m1)_B$}
                    child {
                        node {}
                    }
                }
                
                ;
         \node[above = 0.5cm of N10,align=center,anchor=center] (A12) {$B_{11}$};
         
         \node (N11) [right= 1.7 cm of N10] {$ed(m1)_C$}
                child {
                    node {$mus(m1)_C$}
                    child {
                        node {}
                    }
                }
                child {
                    node {$spa(m1)_E$}
                };

         \node[above = 0.5cm of N11,align=center,anchor=center] (A12) {$C_{11}$};
         
         \node (N12) at (-3.5,-3.5) {$\neg ed(m1)_D$}  
                child {
                    node {$mus(m1)_D$}
                    child {
                        node {}
                    }
                }
                child {
                    node {$am(m1)_E$}
                };
         \node[above = 0.5cm of N12,align=center,anchor=center] (A12) {$D_{11}$};
         
          \node (N10) [right= 3.5 cm of N12] {$am(m1)_E$}
          child {
                node {$spa(m1)_E$}  
                child {
                    node {$mus(m1)_E$}
                    child {
                        node {}
                    }
                } 
                child {
                    node {$hv(m1)_E$}
                    child {
                        node {}
                    }
                } 
                child {
                    node {$pt(m1)_E$}
                    child {
                        node {}
                    }
                }
                child {
                    node {$pbc(m1)_E$}
                    child {
                        node {}
                    }
                }
                child {
                    node {$cml(m1)_E$}
                    child {
                        node {}
                    }
                }
                child {
                    node {$\neg ha(m1)_E$}
                    child {
                        node {}
                    }
                }
                }
                ;
                
         \node[above = 0.5cm of N10,align=center,anchor=center] (A12) {$E_{11}$};

    \end{tikzpicture}
    \caption{Support Relation for query context $\alpha$.}
    \label{fig:args_ex1}
\end{figure*}
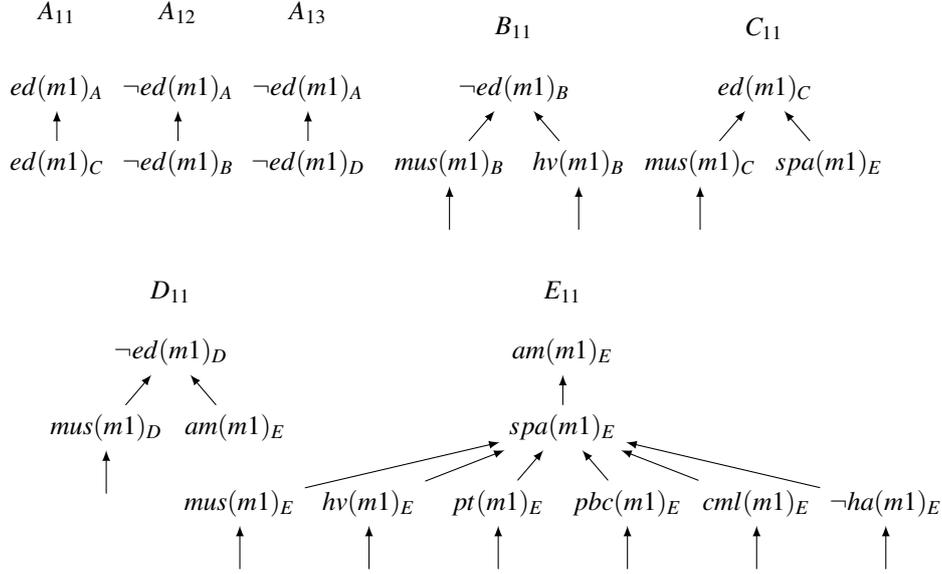

The \textit{Support Relation} ($SR_\alpha$) for the query described in Section \ref{sec:scenario} (Alice's query about the \textit{springtime amanita}) is presented in Figure \ref{fig:args_ex1}. Note that the literals' names were abbreviated to optimise space (e.g. \textit{mushroom(m1)} is represented as \textit{mus(m1)}). The leaf nodes that have the value $\top$ (truth/tautology) are also hidden.

\tikzstyle{level 1}=[level distance=1cm, sibling distance=1.7cm]
\tikzstyle{level 2}=[level distance=1cm, sibling distance=1.7cm]
\tikzstyle{level 3}=[level distance=1cm, sibling distance=1.7cm]

\tikzstyle{bag} = [text width=4em, edge from parent/.style={draw,latex-}]

Note that from the schematic mapping rule $r^{m}_{A3}$ the argument $A_{11}$ is formed, given that there is an argument with $ed(m1)$ as a conclusion ($C_{11}$). Similarly, $A_{12}$ and $A_{13}$ are formed by rule $r^{m}_{A4}$, given that there are two different arguments in different agents that have $\neg ed(m1)$ as a conclusion ($B_{11}$ and $D_{11}$). Argument $B_{11}$ and $E_{11}$ are simply originated from their local rules. Note that the literals originated from the focus rules received in the query are considered as literals defined by each agent (e.g. $mus(m1)_B$ and $hv(m1)_B$ in $B_{11}$; and $mus(m1)_E$, $hv(m1)_E$, etc in $E_{11}$). $C_{11}$ and $D_{11}$, similarly to $A_{11}$ to $A_{13}$, are based on schematic mapping rules, having their leafs associated with conclusions of argument $E_{11}$ ($spa(m1)_E$ and $am(m1)_E$).

From this \textit{Support Relation}, the following steps create the set of justified arguments: (1) the subarguments of $B_{11}$ with heads $mus(m1)_B$ and $hv(m1)_B$ are justified, since their respective rules have empty bodies and there is no argument that defeats them; the same occurs with the subargument with head $mus(m1)_C$ in $C_{11}$ ($C_{11}'$), $mus(m1)_D$ in $D_{11}$ ($D_{11}')$ and $mus(m1)_E$ to $\neg ha(m1)_E$ in $E_{11}$; (2) argument $B_{11}$ is justified, since their subarguments are justified and there is no argument in agent $B$ that defeats $B_{11}$; similarly, argument $E_{11}'$ (with head $spa(m1)_E$) is justified; (3) $C_{11}$ is justified, since one of its two subarguments are justified ($C_{11}'$) and there is an argument ($E_{11}'$) that supports $C_{11}''$; also, $E_{11}$ is justified, given that its only subargument ($E_{11}')$ is justified; $A_{12}$ is not justified because, although it is supported by $B_{11}$ (which is justified), it is defeated by $A_{11}$, since $A_{11}$ is stronger than $A_{12}$ considering the preference order $T_A$ (which states that Catherine's knowledge is preferred to Bob's knowledge);(4) $D_{11}$ is justified; $A_{11}$ is also justified, because there is a justified argument that supports it ($C_{11}$) and, although $A_{11}$ is attacked by both $A_{12}$ and $A_{13}$, it defeats both of them, since Catherine's knowledge is preferred over both Bob and Dennis; (5) similarly to $A_{12}$, $A_{13}$ is not justified (though supported by $D_{11}$) because $A_{11}$ defeats it. $A_{12}$ and $A_{13}$ are rejected because they are defeated by justified arguments. Therefore, $ed(m1)_E$ is taken as a logical consequence of the system.

Let's also suppose that, at the same time, Bob finds a mushroom that he knows it is a \textit{Death Cap}, but he can't remember whether this type of mushroom is edible or not. He then sends the following query to Alice:

\begin{equation*}
\begin{split}
C_Q: \beta \\
p: edible(m2)? \\
r^\mathcal{F}_{\beta_{11}}:  mushroom(m2) \Leftarrow \\
r^\mathcal{F}_{\beta_{12}}:  death\_cap(m2) \Leftarrow
\end{split}
\end{equation*}

\begin{figure}
\centering
    \begin{tikzpicture}[ edge from parent/.style={draw,latex-}]
        \node (N1) at (-5,0) {$\neg ed(m2)_A$}
                        child {
                            node {$mus(m1)_A$}
                            child {
                                node {}
                            }
                        }
                        child {
                            node {$dc(m1)_A$}
                            child {
                                node {}
                            }
                        };
                
        \node[above = 0.7cm of N1,align=center,anchor=center](A1) {$A_{1}$};

      \node (N2)  [right= 1.0cm of N1] {$dc(m2)_B$}  
                child {
                    node {$dc(m2)_A$}
                };
                
         \node[above = 0.7cm of N2,align=center,anchor=center] (B1) {$B_{1}$};

    \end{tikzpicture}
    \caption{Support Relation for query context $\beta$.}
    \label{fig:args_ex1_2}
\end{figure}
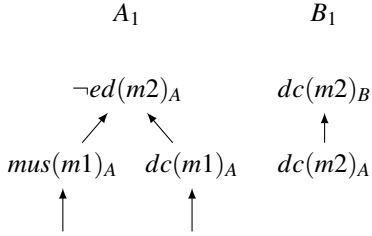

In this case, we have a new \textit{query context} originated from Bob, with different focus rules. From this \textit{query context} a different \textit{Support Relation} is created, which will be called $SR_{\beta}$, and the proof trees for it are very different from $SR_\alpha$, as shown in Figure \ref{fig:args_ex1_2}. Note that, again, the literal names were abbreviated (e.g. $death\_cap(m2)$ is written as \textit{dc(m2)}).

It is interesting to note that both Support Relations are very different from each other, considering that most of the rules used are the same for both queries, and also that they not interfere in each other. This exemplifies how different \textit{foci} or contexts lead to different argumentation frameworks in the reasoning process.

\section{\uppercase{Implementation}}
\label{sec:implementation}
An initial multi-agent implementation was made using the framework Jason \cite{jason}, which uses the language \textit{AgentSpeak} for programming agents based on the BDI (\textit{Belief-Desire-Intention}) model. 

To optimise space and simplify the understanding of the algorithm, we present in this paper a simplified  version of the program in \textit{AgentSpeak}, highlighting its main points. 

Agent's rules are encoded using the following predicate style:

\begin{lstlisting}[caption={Example of rule encoded in \textit{AgentSpeak}.},label={lst:rules}]
mapping_rule(l12, hunterA, ~edible(M)[source(hunterA)], [death_cap(M)[source(any)]]).
...
\end{lstlisting}

The example above shows how rule $r^m_{A1}$ is encoded. Note that the literals in the head and body are annotated with a \texttt{source} predicate, which indicates the agent that defines the literal. In the case of  $death\_cap(M)$, the source is marked as \texttt{any}, indicating that it is a schematic literal that may be defined by any agent.

When Alice queries about $edible(m1)$, she first tries to answer it by herself by adding the goal \texttt{!query(q1, edible(m1), hist(q1, edible(m1), [edible(m1)]), Rf)} to the agent's mental state, where $q1$ is the query context identification (which is generated automatically), \texttt{edible(m1)} is the literal that we want to find the truth value for, and \texttt{hist(q1, edible(m1), [edible(m1)])} the initial history for the query, which is used to detect cycles in the global knowledge base (the union of all the belief bases), and \texttt{Rf} refers to the set of current focus rules of Alice. When this goal is added to Alice, one of the following plans is activated:

\begin{lstlisting}[caption={Action plans that represent the options when a query is received.},label={lst:rules}]
+!query(CId,P,Hist,Rf)[source(A0)]: locally(P) <-
    +query_context(CId,A0);
    !return_to_caller(CId,P).
    
+!query(CId,P,Hist,Rf)[source(A0)]: locally(~P) <-
    +query_context(CId,A0);
    !return_to_caller(CId,P).

+!query(Cid,P,Hist,Rf)[source(A0)]: not(locally(P)) & not(locally(~P)) 
        & not(support_finished(P)) <-
  +query_context(CId,A0);
  !add_focus_rules(CId,Rf);
  !seek_support(CId,P);
  !seek_support(CId,~P).
\end{lstlisting}

The first plan is executed when a local conclusion is found (\texttt{locally(P)}). The details of the rules involved in the calculation of the \texttt{locally(P)} predicate will not be covered, since they are simply a reproduction of the well known logic programming reasoning. If a local conclusion for the complement of $P$ is found (\texttt{$\sim$P}), then the second action is executed. In both cases, the goal \texttt{return\_to\_caller(CId,P)} is added in order to return an answer. A belief \texttt{query\_context(Cid, A0)} is also added to record which agent issued the query (\texttt{A0}). This is very important later when returning an answer for the query.

When neither \texttt{P} nor \texttt{$\sim$P} are found to be locally provable, the third action plan is executed. In this case, a new goal \texttt{!add\_focus\_rules(CId,Rf)} is added in order to add the focus rules (\texttt{Rf}) tagged with the query context identification \texttt{CId} to the agent.

\begin{lstlisting}[caption={Action plan to add received focus rules as local rules.},label={lst:focus_rules}]
+!add_focus_rules(CId,Rf): true <-
  for (.member(L,Rf)) {
    if (not(rule(_,_L,_)[context(CId)])) {
      ?rule_id(IDN);
      .concat("tr",IDN,ID);
      -+rule_id(IDN+1);
      +rule(ID,L,[])[context(CId)];   
    }
  }.    
\end{lstlisting}

Note that each rule created from the focus rules are annotated with a \texttt{context(CId)} clause, which indicates that the rule is bound to this particular query context.

Other two goals \texttt{!seek\_support(CId,P)} and \texttt{!seek\_support(CId,$\sim$P)} are added to try to find a support for the literals \texttt{P} and \texttt{$\sim$P} in question. Basically, the support is found by iterating over all rules whose head is \texttt{P} (or \texttt{$\sim$P}), then performing a new query for each member of the body of these rules:

\begin{lstlisting}[caption={Action plans responsible for finding support for a rule},label={lst:support}]
+!seek_support(CId,P,Hist): rule(_,_,P,_)[context(C)] & (C == CId | C == any) <-
	for (rule(R,_,P,Body)[context(C)]){
        +waiting_for_support_return(CId,R,P,Body,[],[],1)
        for (.member(B[source(Ag)], Body)){
            !evaluate_body_member(CId,R,B,Ag,Hist) } }.
(...)

+!evaluate_body_member(CId,R,B,Ag,Hist): 
		Hist = hist(CId,P,L) & not(.member(B,L)) 
		& Ag == any <-	
	!ask_all_known_agents(CId,R,B,hist(CId,B,[B|L])).
	
+!evaluate_body_member(CId,R,B,Ag,Hist): 
		Hist = hist(CId,P,L) & not(.member(B,L))
		& Ag \== any<-
	.send(Ag, achieve, query(CId,B,hist(CId,B,[B|L]))).
	
+!evaluate_body_member(CId,R,B,Ag,Hist):
		Hist = hist(CId,P,L) & .member(B,L)
		& .my_name(Me) & Ag \== Me <-
	+cycle(CId,R,B);
	(...).
    
\end{lstlisting}


 Note that, for each body member of each rule with head \texttt{P}, the agent checks whether its source is ``\texttt{any}'' or refers to a specific agent. If it is the former, then a goal  \texttt{ask\_all\_known\linebreak\_agents(CId,R,B,hist(CId,B,[B|L]))} is added, which simply sends the goal \texttt{query(CId,B,hist(CId,B,[B|L]))} (where \texttt{B|L} is the addition of \texttt{B} to the history of the new query) to every known agent. Otherwise, it sends the query only to the agent that it knows that defines \texttt{B}. Note that both the second and the third plans are triggered only if \texttt{B} is not in \texttt{Hist}. If  \texttt{B} is in \texttt{Hist}, then we have a cycle in the global knowledge base, which then activates the fourth plan. This plan do not issue any new query for \texttt{B} and a \texttt{cycle(CId,R,B)} belief is added to indicate that the rule \texttt{R} cannot be used to support \texttt{P}, although it will be considered an \textit{unblocked rule} if none of the other body members is considered \textit{false}, as shown in Listing \ref{lst:support}. 


When an agent receives an answer for a query to a literal \texttt{B}, it receives a belief \texttt{answer(B,TV,BS,SS,Ag)}, where \texttt{TV} can be one of the three possible truth values: \textit{true}, \textit{false} or \textit{undefined}. The code for the action plans that are activated when these answers are received will be omitted because of the space constraint. \texttt{BS} and \texttt{SS} are respectively the \textit{Blocking Set} and the \textit{Supportive Set} for the literal. The BS is the set of literals in the body of an \textit{unblocked rule} that supports \texttt{P}, whereas the SS is the set of literals in the body of an \textit{applicable rule} for \texttt{P}. An \textit{unblocked rule} is a rule with head \texttt{P} such that all literals in its body have \textit{true} or \textit{undefined} as truth value. An \textit{undefined} truth value indicates that a cycle was found during the processing of the query. An \textit{applicable rule} is a rule with head \texttt{P} such that all literals in its body have \textit{true} as truth value. All applicable rules are also unblocked rules, but the opposite is not true. According to the semantics presented by \cite{bikakis2010}, argumentation lines where cycles occur cannot be used to support a literal (to be \textit{true}), but it also may not imply that the literal is rejected (\textit{false}). 

When all literals in the body of a rule that supports \texttt{P} have an answer, then the following action plan is triggered. Note that this action plan is triggered only if no other rule was found to be \textit{unblocked} yet, or if the BS of the new rule (\texttt{BSr}) evaluated is stronger than the BS of the previous \textit{unblocked} rule found (\texttt{BSp}). The BS of a rule is composed of the members of the rule's body if they are foreign literals. If a body member \texttt{B} is a local literal, then the elements of its BS is included in the BS of the rule (\texttt{BSr}). The function \texttt{stronger} performs the role of the \textit{Str} function defined in Section \ref{sec:arg_construction}, comparing the strength based on the literals pertaining to each BS.


\begin{lstlisting} [caption={Action plan triggered after every body member of a rule is evaluated.},label={lst:waiting1}]
+waiting_for_support_return(CId,R,P,[],BSr,SSr,1):
		rule(R,_,P,Body)[context(C)] & (C == CId | C == any)
		& (not(unblocked(CId,P,BSp)) | stronger(BSr,BSp,T,BSr)) <-
	-+unblocked(CId,P,BSr);
	if (not(cycle(CId,R,_))) {
		if (not(supported(CId,P,SSp)) | stronger(SSr,SSp,T,SSr)) {
			-+supported(CId,P,SSr); } }
	-+waiting_for_support_return(CId,R,P,[],BSr,SSr,2).
\end{lstlisting}

Note that the literal is considered supported iff a cycle was not found for the rule (\texttt{R}), because a cycle indicates that \texttt{P} is only unblocked, and not supported, by \texttt{R}.




There are other action plans to synchronise the results of each \texttt{seek\_support} goal added to the agent, but they will be omitted because of space constraints. The following action plans are called at the end of the processing of a query, when both the \texttt{support} for \texttt{P} and \texttt{$\sim$P} are finished or if the answer was found locally.


\begin{lstlisting} [caption={Action plans responsible for returning a response for the query},label={lst:response}]
+!return_to_caller(CId,A0,P): 
		(locally(CId,P) |
			(supported(CId,P,SSp)
				& (not(unblocked(CId,~P,BSq)) | stronger(SSp,BSq,T,SSp)))) <-
	.send(A0, tell, answer(P,true,BS,SS,Me)[context(CId)]);
	!clear(CId, A0, P).
 
+!return_to_caller(CId,A0,P): 
		(locally(CId, ~P) |
			not(unblocked(CId,P,SSp)) |
			(( not(supported(CId,P,SSp))
					| (unblocked(CId,~P,BSq)  & stronger(BSq,SSp,T,BSq)))
				& supported(CId,~P,SSq)
				& stronger(SSq,BSp,T,SSq)))
		<-
    .send(A0, tell, answer(P,false,BS,SS,Me)[context(CId)]);
    !clear(CId, A0, P).
    
-!return_to_caller(CId,A0,P): 
		support_finished(CId,P,BSp,SSp)  <-
    .send(A0, tell, answer(P,undefined,BS,SS,Me)[context(CId)]);
    !clear(CId, A0, P).
\end{lstlisting}

In summary, a \textit{true} answer for \texttt{P} is returned if it is proved locally, or if \texttt{P} is supported (an applicable rule was found) and \texttt{$\sim$P} is either not even unblocked or it is unblocked but the SS of \texttt{P} is stronger than the BS of \texttt{$\sim$P}. The answer will be \textit{false} if \texttt{$\sim$P} is proved locally, or if the conditions for answering \textit{true} do not hold and, additionally, if \texttt{$\sim$P} is supported and the SS of \texttt{$\sim$P} (\texttt{SSq}) is stronger than the BS of  \texttt{P} (\texttt{BSp}). Otherwise, its answer is \textit{undefined}.



\begin{figure*} [hbt] 
\centering
\caption{Sequence diagram for the reasoning process of the mushroon hunters scenario.}
\includegraphics[width=0.9\textwidth]{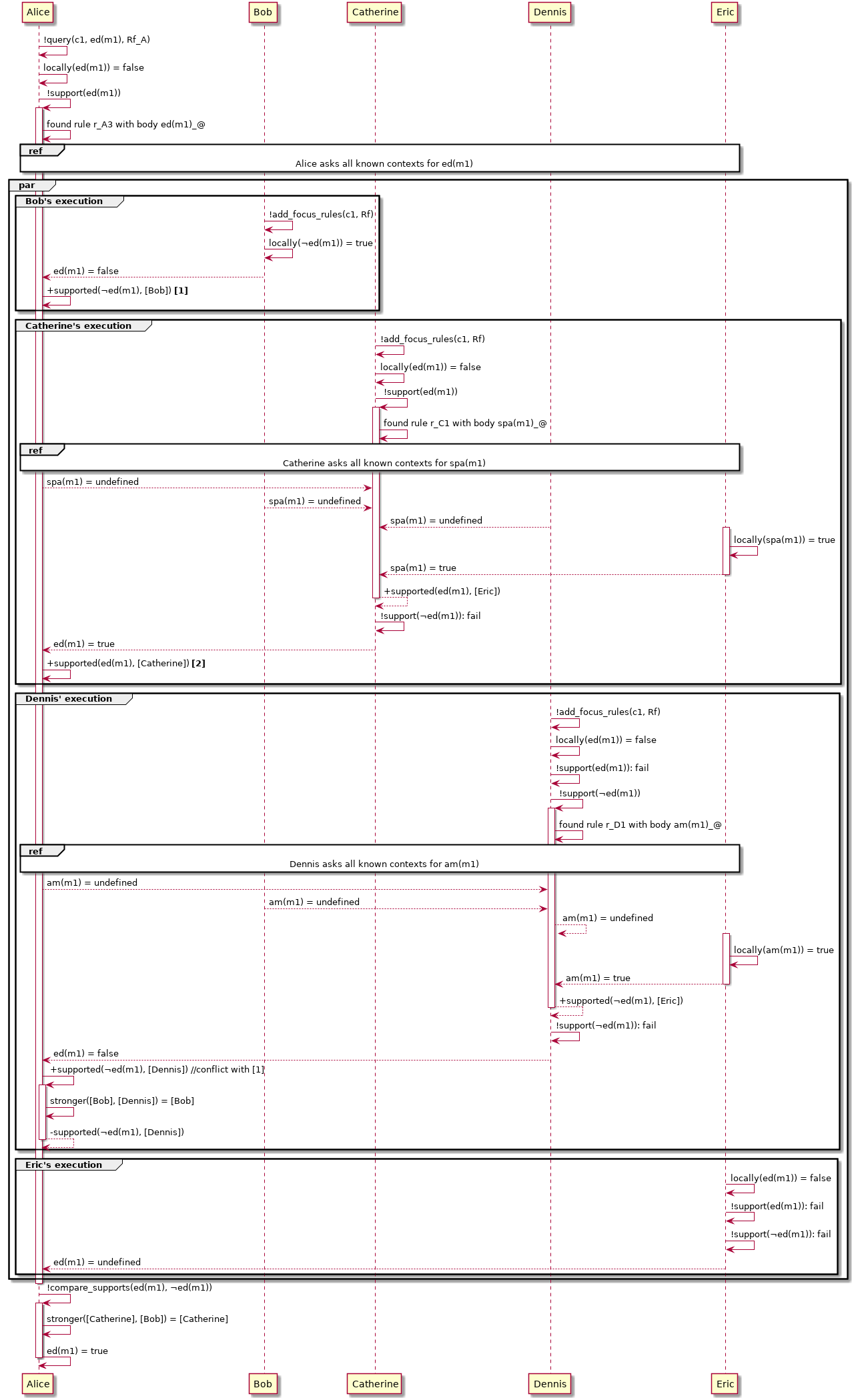}
\label{fig:seqdiagram} 
\end{figure*}

Figure \ref{fig:seqdiagram} shows a sequence diagram which illustrates the execution of the scenario presented in Section \ref{sec:scenario}.

\section{\uppercase{Related Works and Conclusion}}
\label{sec:conclusions}

The main related work is the one by \cite{bikakis2010}, which the model presented in this works extends. It enables decentralised distributed reasoning based on a distributed knowledge base, such that the knowledge from different knowledge bases (called \textit{contexts}) may conflict with each others. Such conflicts are solved by means of defeasible logic considering the existence of mapping rules, i.e. rules that have premises defined by other knowledge bases. They also present an algorithm for distributed query evaluation based on the defined semantics. However, their work still lacks some generality, in the sense that there exists many use case scenarios that are not possible to represent in their system. One kind of such scenarios are the ones that require that agents share relevant knowledge when issuing a query to others. Another kind of scenarios are those in which the bindings among the agents (defined by means of mapping rules) are not static, such as in knowledge-intensive and dynamic environments.

Other works with similar goals and approach are the peer-to-peer inference systems, such as the proposed by \cite{adjiman2006}, which also provides decentralised distributed reasoning based on distributed knowledge, but do not provide ways to resolve conflicts. Such systems also do not deal with the idea of sharing relevant context knowledge when issuing a query. Another relevant work is the distributed argumentation with defeasible logic programming by \cite{thimm2008}. However, its reasoning is not fully decentralised, depending on a moderator agent that has part of the responsibility of constructing arguments. 

Therefore, our framework extends the contextual defeasible logic of \cite{bikakis2010} as a multi-agent model including the idea of focus and its effects in distributed reasoning, as well as giving a glimpse of how CDL can be adapted to enable its use in dynamic and knowledge-intensive environments where agents come in and out of the environment at random times and have their beliefs constantly updated and revised. Furthermore, we present a multi-agent implementation based on the BDI agent architecture.

Our future work includes exploring the possibility of using partial order instead of total order in scenarios where there may be incomparable agents, as suggested in \cite{gottifredi2018}. Another relevant future improvement for this work is creating mechanisms to allow agents to internalize conclusions or arguments from previous queries when they are reliable enough, and to not do it if the information used was not so reliable. This would avoid repeating the distruted query processing for the same literal if the arguments used to support it are reliable enough. Other ideas include developing new conflict resolution strategies and optimizations for cases where the focus rule set is too big to share among agents.

\bibliographystyle{apalike}
{\small
\bibliography{example}}

\begin{thebibliography}{}

\bibitem[Adjiman et~al., 2006]{adjiman2006}
Adjiman, P., Chatalic, P., Goasdou{\'e}, F., Rousset, M.-C., and Simon, L.
  (2006).
\newblock Distributed reasoning in a peer-to-peer setting: Application to the
  semantic web.
\newblock {\em Journal of Artificial Intelligence Research}, 25:269--314.

\bibitem[Benerecetti et~al., 2000]{benerecetti2000}
Benerecetti, M., Bouquet, P., and Ghidini, C. (2000).
\newblock Contextual reasoning distilled.
\newblock {\em Journal of Experimental \& Theoretical Artificial Intelligence},
  12(3):279--305.

\bibitem[Bikakis, 2009]{bikakis2009tese}
Bikakis, A. (2009).
\newblock {\em Defeasible Contextual Reasoning in Ambient Intelligence}.
\newblock PhD thesis, Computer Science Department, University of Crete.

\bibitem[Bikakis and Antoniou, 2010]{bikakis2010}
Bikakis, A. and Antoniou, G. (2010).
\newblock Defeasible contextual reasoning with arguments in ambient
  intelligence.
\newblock {\em IEEE Transactions on Knowledge and Data Engineering},
  22(11):1492--1506.

\bibitem[Bordini et~al., 2007]{jason}
Bordini, R.~H., H{\"u}bner, J.~F., and Wooldridge, M. (2007).
\newblock {\em Programming multi-agent systems in AgentSpeak using Jason},
  volume~8.
\newblock John Wiley \& Sons.

\bibitem[Gottifredi et~al., 2018]{gottifredi2018}
Gottifredi, S., Tamargo, L.~H., García, A.~J., and Simari, G.~R. (2018).
\newblock Arguing about informant credibility in open multi-agent systems.
\newblock {\em Artificial Intelligence}, 259:91 -- 109.

\bibitem[Homola et~al., 2015]{homola2015}
Homola, M., Patkos, T., Flouris, G., {\v{S}}efr{\'{a}}nek, J., {\v{S}}imko, A.,
  Frt{\'{u}}s, J., Zografistou, D., and Bal{\'{a}}{\v{z}}, M. (2015).
\newblock {Resolving conflicts in knowledge for ambient intelligence}.
\newblock {\em Knowledge Engineering Review}, 30(5):455--513.

\bibitem[Thimm and Kern-Isberner, 2008]{thimm2008}
Thimm, M. and Kern-Isberner, G. (2008).
\newblock A distributed argumentation framework using defeasible logic
  programming.
\newblock In {\em Proceedings of the 2008 Conference on Computational Models of
  Argument: Proceedings of COMMA 2008}, pages 381--392, Amsterdam, The
  Netherlands, The Netherlands. IOS Press.

\end{thebibliography}



\end{document}